\def\year{2022}\relax
\documentclass[letterpaper]{article} % DO NOT CHANGE THIS
\usepackage{aaai22}  % DO NOT CHANGE THIS
\usepackage{times}  % DO NOT CHANGE THIS
\usepackage{helvet}  % DO NOT CHANGE THIS
\usepackage{courier}  % DO NOT CHANGE THIS
\usepackage[hyphens]{url}  % DO NOT CHANGE THIS
\usepackage{graphicx} % DO NOT CHANGE THIS
\usepackage{natbib}  % DO NOT CHANGE THIS AND DO NOT ADD ANY OPTIONS TO IT
\usepackage{caption} % DO NOT CHANGE THIS AND DO NOT ADD ANY OPTIONS TO IT
\DeclareCaptionStyle{ruled}{labelfont=normalfont,labelsep=colon,strut=off} % DO NOT CHANGE THIS
\usepackage{algorithm}
\usepackage{algorithmic}

\usepackage{newfloat}
\usepackage{listings}
\floatstyle{ruled}
\newfloat{listing}{tb}{lst}{}
\floatname{listing}{Listing}
\usepackage[disable]{todonotes}
\usepackage{soul}
\usepackage{xurl}
\usepackage{array}
\newcommand{\OurTitle}{Enabling NAS with Automated Super-Network Generation}

\title{\OurTitle}
\author{
    J. Pablo Mu\~{n}oz\textsuperscript{\rm 1}, %\thanks{With help from the AAAI Publications Committee.},
    Nikolay Lyalyushkin\textsuperscript{\rm 2},
    Yash Akhauri\textsuperscript{\rm 1},
    Anastasia Senina\textsuperscript{\rm 2},
    Alexander Kozlov\textsuperscript{\rm 2},
    Nilesh Jain\textsuperscript{\rm 1}
}
 \affiliations{
     \textsuperscript{\rm 1}Intel Labs,  
     \textsuperscript{\rm 2}Intel Corporation
    \\%Association for the Advancement of Artificial Intelligence\\
     \texttt{\{pablo.munoz, nikolay.lyalyushkin, yash.akhauri, anastasia.senina,} \\\texttt{alexander.kozlov, nilesh.jain\} @intel.com}
}
\begin{document}
\maketitle
\begin{abstract}

Recent 
Neural Architecture Search (NAS) solutions have produced impressive results training super-networks and then deriving subnetworks, a.k.a. child models %that are derived from a super-network 
that outperform expert-crafted models from a pre-defined search space. Efficient and robust subnetworks can be selected for resource-constrained edge devices, allowing them to perform well in the wild. 
However, constructing super-networks for arbitrary architectures is still a challenge that often prevents the adoption of these approaches. To address this challenge, we present BootstrapNAS, a software framework for automatic generation of super-networks for %one-shot 
NAS. BootstrapNAS takes a pre-trained model from a popular architecture, e.g., ResNet-50, or from a valid custom design, and automatically creates a super-network out of it, then uses state-of-the-art NAS techniques to train the super-network, resulting in subnetworks %, a.k.a. child models, 
that significantly outperform the given pre-trained model. We demonstrate the solution by generating super-networks from arbitrary model repositories and make available the resulting super-networks for reproducibility of the results. 

\end{abstract}

\section{Introduction}

The great variety of edge devices in which \emph{Deep Learning} models might be deployed has motivated the development of solutions for optimizing these models and improving their performance on a selected device. A successful approach has been to use Neural Architecture Search (NAS) to discover efficient and robust models that can be deployed in the wild on a particular edge device. 

Early NAS solutions trained each candidate architecture from scratch, taking a significant amount of time to produce decent results. \emph{Weight sharing} has allowed more efficient NAS approaches that maintain a single structure, i.e., a super-network, sometimes referred to as the \emph{one-shot} \cite{Bender2018UnderstandingAS, liu2018darts, pham2018efficient, cai2018proxylessnas, xie2020snas, guo2020single}, \emph{single-stage} \cite{Bignas}, or \emph{once-for-all} \cite{cai2020once} network depending on other properties of each NAS approach. A few of these 
approaches are hardware-aware, for instance, by incorporating the target device's latency measurements, but are limited to a single target device 
, e.g., \cite{cai2018proxylessnas}, %fbnetwu2018,, BermanAowsAdaptive}, %and 
having to run again the 
NAS procedure when an optimal model for a new target device is requested. More recently, NAS approaches have been able to decouple train and search stages, enabling a single training session and the repeated search of new derived models for multiple target devices with different hardware configurations \cite{cai2020once, Bignas}. These \emph{once-for-all} or \emph{single-staged} super-networks %have demonstrated 
produce 
large spaces of subnetworks. The weights of the super-network are optimized and exploration of suitable sub-networks is guided by a \textit{search strategy} and a \textit{performance estimation strategy} \cite{ElskenNASSurvey}. In some cases, these approaches produce models that can be immediately deployed, e.g., \cite{cai2018proxylessnas, cai2020once, Bignas}, while in other cases, additional fine-tuning of the candidate subnetworks can yield better accuracy.  
Smaller subnetworks might satisfy the requirements of resource-constrained devices while maintaining the accuracy of bigger subnetworks.  

The
construction of the super-network, and hence the generation of the search space, present several challenges. 
Given that there is no consensus on the optimality of one-shot NAS, driven by doubts on whether super-network optimization aligns with the objective of NAS \cite{zhang2021does}, generating a search space that contains well performing sub-networks needs to be automated effectively. Further, existing search spaces are designed for research endeavors which may not align with tasks for which a neural network has to be deployed. 

Expert practitioners have to construct these super-networks for instance by overparameterizing a well-known architecture, e.g., MobileNet-V3 \cite{MobilenetV3_Howard_2019_ICCV}. However, this is usually a complicated process that prevents an average user %of Deep Learning models 
from further optimizing their existing pre-trained models so they can improve their performance when deployed in the wild. 
%The solution in this paper addresses the challenge of automatically creating super-networks for these kind of approaches, sometimes termed once-for-all networks. 

We present BootstrapNAS, a software framework for automatic generation of NAS super-networks. BootstrapNAS is implemented in the Neural Network Compression Framework (NNCF) \cite{kozlov2020neural}. NNCF works with PyTorch and TensorFlow, and supports a wide range of compression algorithms such as quantization and pruning. We utilize NNCF's graph tracing and analysis capabilities to enable the BootstrapNAS solution.

Our contributions can be summarized as follows:
\begin{itemize}
    \item A software framework with a set of methods for automated super-network generation. 
    \item The application of state-of-the-art methods for training the automatically generated super-network. %n example of the application of a state-of-the-art One-Shot NAS method built on top of the automatic super-network generation.
    \item Demonstration of the feasibility of the proposed methods and preliminary results on two examples of super-networks. %Results of experiments that demonstrate the feasibility of the proposed method for generating and training super-networks.
\end{itemize}

\section{Automated Generation of Super-Networks}% for One-Shot NAS}
% \begin{wraptable}{r}{0.35\linewidth} % wrapFig is not allowed 
\begin{table}[h]
  \caption{Notation}
  \label{table:notation}%
%   \centering
  {
        \begin{tabular}{|m{1em} m{9em}||m{0.8em} m{8em}|}
    \hline
    $\Omega$      & Super-network                      & $L^{\Omega}$ 
    & Set of layers of $\Omega$  \\
    $a_i$        & Subnetwork $i$                     & $l^{\Omega}_i$ %sn}_i$
    & Layer $i$ of $\Omega$ \\
    $a_{min}$     & Minimal subnetwork & $L^{i}$     & Set of layers of $a_i$\\
    $a_{max}$   & Maximal subnetwork & $l^{i}_j$   & Layer $j$ of $a_i$\\
    $m$        & Pre-trained model & $L^{i}_s$   & Set of \emph{static} layers \\ 
    $A$          & Set of all subnetworks             &    & of $a_i$ \\  
    $A_o$        & Set of Pareto-optimal & $L^{i}_e$ & Set of \emph{elastic} layers \\ 
    & subnetworks & & of $a_i$\\
    \hline
    \end{tabular}%
    }
% }\vspace{-10mm}
\end{table}
% \end{wraptable}

\paragraph{Super-network.} A super-network, $\Omega$, is a neural network that is composed of a set of layers, $L^{\Omega}$, which for our purposes we divide into two subsets, $L^{\Omega}_e$ (elastic layers) and $L^{\Omega}_s$ (static layers), i.e., $L^{\Omega} = L^{\Omega}_e \cup L^{\Omega}_s$, and a set of weights, $W$, associated with those layers. Notice that in our formulation, we start by considering a super-network, as a typical neural network, and it is through BootstrapNAS' top-down approach that \emph{elasticity} (defined below) is automatically added, and which allows the later derivation of subnetworks.

\paragraph{Subnetwork.} A subnetwork or child model, $a_i$, is a neural network that shares some (or all) of the elements of the super-network, $\Omega$, s.t., if $L^i$ is the set of the layers in $a_i$, then $L^{i} \subseteq L^{\Omega}$. Notice that $L^{i}$ is also composed of the two types of layers introduced above, i.e., $L^i = L^i_s \cup L^i_e$, and $\forall l (l \in L^i_s \Leftrightarrow l \in L^{\Omega}_s)$.

\paragraph{Static Layers.} We refer to $L^{i}_s$ (or $L^{\Omega}_s$ for that matter) as \emph{static} layers, and, as we will discuss below, they have a fixed configuration for all the subnetworks and the super-network.

\paragraph{Elastic Layers.} $L^{i}_e$ are those layers that can have variable values in their properties. For instance, in the case of a convolution layer, they might have variable values for its \emph{width} (number of channels) or \emph{kernel size}, e.g., a layer with $x$ or $y$ number of channels, and $z$x$z$ kernels., e.g., 7x7. We refer to $L^{i}_e$ as \emph{elastic} layers.

We denote $A$ to be the set of all the subnetworks (which includes the super-network). The literature uses the term \emph{elasticity} to describe the property of layers that can vary their configurations, e.g., layer $j$ in subnetwork $a_i$ i.e., $l^i_j$, might have $x$ number of \emph{active} channels, while another subnetwork,  $a_k$, %$l$ in layer $j$, i.e., $l^i_j$,
might have $y$ number of \emph{active} channels for the same layer, $l^k_j$. Although the same layer is present in both subnetworks, their layer configurations might be different. Thus, subnetworks are partitions of the super-network, and have different values for the properties of their layers, e.g., width or kernel size in the case of convolution layers, unless the particular subnetwork in consideration is also the super-network.

\paragraph{Pre-trained Model to Super-Network.} BootstrapNAS uses NNCF's capabilities to trace a given pre-trained model, $m$, and convert it into a super-network, $\Omega$, s.t., if $L^m$ are the layers in $m$, $\forall l (l \in L^{m} \Leftrightarrow l \in L^{\Omega})$ and $W^{\Omega} = W^{m}$. The conversion procedure must guarantee that both models, that is, the pre-trained model and the super-network, will produce similar results on a dataset, $D_{val}$, i.e., $Cost(m, D_{val}) \cong Cost(\Omega, D_{val})$. This is validated by BootstrapNAS after the super-network has been generated. Before conversion of a pre-trained model to a super-network all layers are static, i.e., $L^{\Omega} = L^{\Omega}_s$. BootstrapNAS detects layers that can be made elastic, e.g., a convolution layer, by checking the type of the underlying operation in the layer in consideration and comparing it to the supported operations that can be made elastic. The selected static layers becomes elastic without rewriting the model's code by injecting a mechanism that can capture inputs and parameters before the layer's execution, apply transformations on the layer's tensors and run the underlying operation with the modified parameters and inputs. %NNCF provides this functionality and  BootstrapNAS enables the elastic features of a layer, by inserting transformation that dynamically varies options for the number of channels or kernel sizes in the case of convolution operations. All this functionality is added before the super-network training stage.

In addition to varying its width and internal layer properties, a subnetwork, $a_i$, can be shallower than its parent super-network, so it is possible that $|L^{i}|$ may be less than $|L^{\Omega}|$. The change in depth is accomplished by \emph{omitting} individual layers or groups (blocks) from the super-network to derive subnetwork $a_i$. BootstrapNAS accomplishes this omission of layers by temporarily removing them from the computational graph if the subnetwork is selected during training. During the conversion of the pre-trained model to a super-network, BootstrapNAS automatically detects layers or blocks that can be skipped by analyzing groups of layers, and determining whether they could be skipped/removed without creating inconsistencies between the output tensors of the previous block and the input dimensions of the following block. BootstrapNAS checks for inconsistencies that might occur when $activating$ certain number of channels in a layer, since this number has to be consistent with its adjacent layer(s).

BootstrapNAS automatically generates the NAS search space by creating a configuration of elasticity for each layer. It starts by considering the maximum possible value of a layer's property based on the value of this property on the original pre-trained model, e.g., number of channels for a layer, and then including alternative configurations with smaller values and steps. For instance, if the pre-trained model used 512 channels in a layer, BootstrapNAS can generate alternative configurations, e.g., $\{512, 256, 128\}$ for the possible number of channels in the derived subnetworks. The number of alternatives, stopping criteria and decreasing step is easily configurable. Otherwise, defaults are used. The search space generation also takes into account the blocks that might be skipped to model how subnetworks will vary in depth. A super-network can easily end up deriving billions of possible subnetworks, depending on how many possible configurations BootstrapNAS might allow on each layer.

\paragraph{Minimal and Maximal Subnetworks.} A subnetwork $a_i$ is considered to be the minimal subnetwork, $a_{min}$, if it configuration uses the minimal possible values for each elastic dimension on each elastic layer. On the contrary, a differente subnetwork $a_i$ is considered to be maximal if it uses the maximal value for each elastic dimension in the each elastic layer. Note that $a_{max}$ is equivalent in its architecture to the given pre-trained model. 

\section{Super-Network Training and Subnetwork Search}

The super-network generated from the pre-trained model is suitable to the application of 
state-of-the-art super-network training techniques. %One-Shot NAS algorithm. A 
For instance, a proven algorithm is \emph{Progressive Shrinking} by  \cite{cai2020once}. As its name suggests, it trains the super-network by allowing the sampling of smaller random subnetworks at each training stage (kernel size, depth, and width), hence increasing the variety of subnetwork configurations. There are other techniques for training super-networks. For instance, instead of focusing in subnetworks of a decreasing size for each stage, BootstrapNAS can apply the ``sandwich'' rule proposed in \cite{yu_slimmable19}, in which, % Using this technique, 
at each batch of data, a few subnetworks are sampled: the minimal subnetwork, $a_{min}$, the maximal subnetwork, $a_{max}$ and other $n$ randomly sampled subnetworks. The gradients are aggregated and the weights of the super-network are updated accordingly.    

Knowledge distillation can also be applied during training.  %\cite{hinton2015distilling}. 
The soft labels from the original pre-trained model, $m$ or from the maximal subnetwork, $a_{max}$ can be used to compute the loss of the sampled subnetworks. Using the soft labels of $a_{max}$ is referred to as \emph{inplace distillation} in the literature \cite{yu_slimmable19}. 

BootstrapNAS' implementation has an \emph{elasticity handler} object that maintains a registry of the possible configurations that a elastic layer (or the set of elastic blocks in the case of depth) might take, allowing for an efficient sampling of subnetworks. When a subnetwork configuration is selected, BootstrapNAS activates the corresponding configuration at each layer, so the forward and backward passes can be done. 

The level of elasticity depends on the size of the given pre-trained model, as well. An overparameterized pre-trained model will allow for the generation of a larger search space. Notice that although an immense number of subnetworks can be derived from the super-network, the space required to store all this information never exceeds the space required to store the super-network, which is a great benefit of weight-sharing approaches. BootstrapNAS' cost to maintain the information of the possible configurations that a layer can have is minuscule in comparison with the size of the model.

As defined above, $A$ is the set of all the possible subnetworks that can be derived from a super-network. Once BootstrapNAS completes the super-network training stage, its next goal is to find $k$ Pareto-optimal subnetworks. That is, BootstrapNAS constructs a set, $A_o \subseteq A$, s.t., $|A_o| = k$. BootstrapNAS currently uses %evolutionary algorithms (\cite{CoelVanVLamo02}), 
the Non-dominated Sorting Genetic algorithm II (NSGA-II) by \cite{nsgaII} %in particular is the
as
default algorithm to search for the set of Pareto-optimal subnetworks. NSGA-II evolves a population and then ranks the various configurations to produce a set of non-dominated solutions.  Although BootstrapNAS currently uses NSGA-II by default, nothings prevents BootstrapNAS to incorporate other search algorithms in a subsequent search.

\begin{figure*}[ht] %H
\centering
\includegraphics[width=0.80\linewidth]{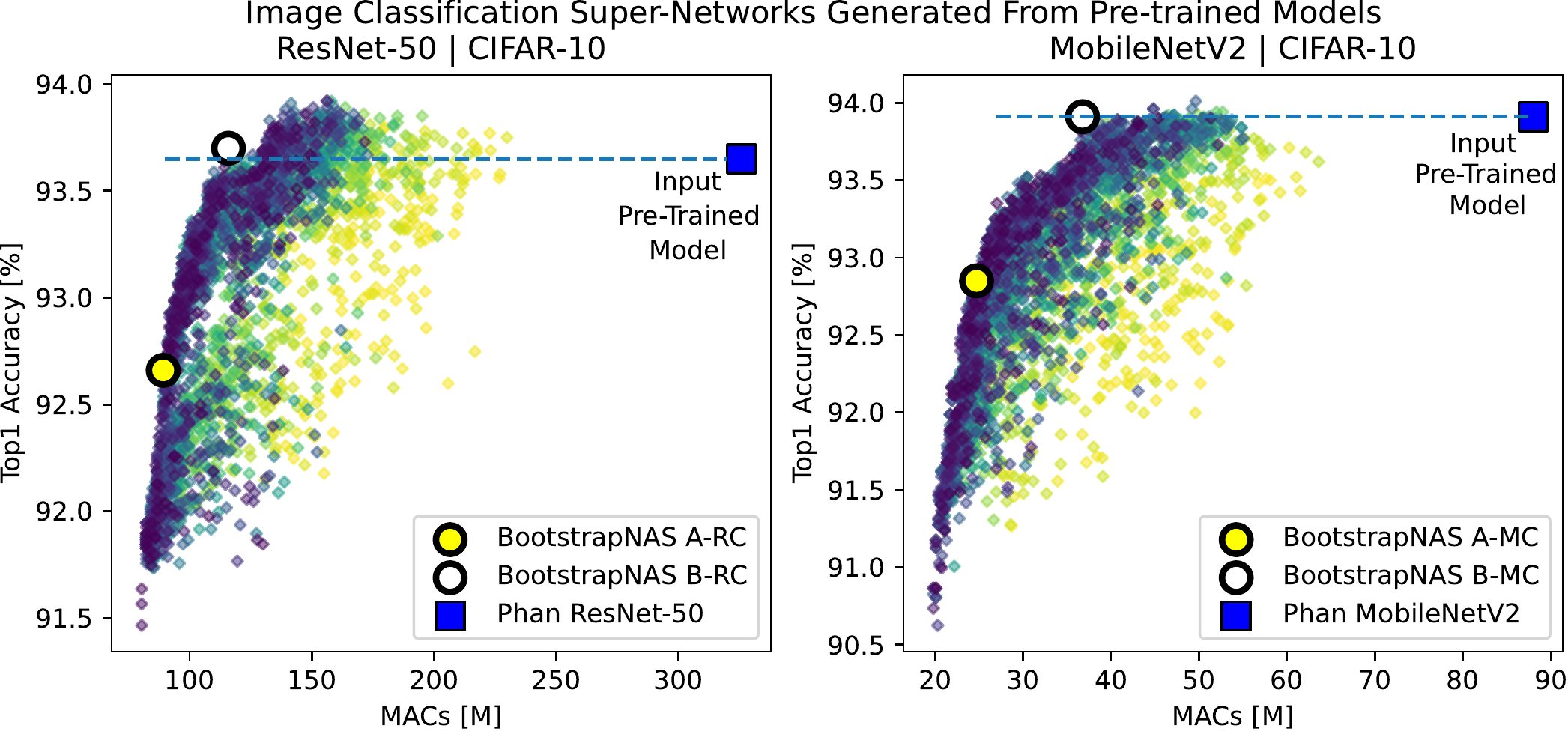}
%\todo[inline]{Replace with updated MBV2}
% \includegraphics[width=\textwidth]{figures/supernet_macs_all.pdf}
\caption{Subnetwork search using NSGA-II on the FP32 model spaces of two super-networks automatically generated by BootstrapNAS from pre-trained models. %for image classification. 
Each plot illustrates the progression of NSGA-II when discovering a Pareto front. The darker marks represent subnetworks evaluated late in the search process. All subnetworks above the dashed lines outperform the pre-trained models given as input, in both objectives, accuracy and MACs. All subnetworks to the left of the input model outperform the input model in MACs. We highlight two subnetworks in addition to the given pre-trained model. 
}
\label{fig:nsga_all}
\end{figure*}

\section{Evaluation of Two Examples of Automatically Generated Super-Networks}
\label{sec:evaluation}

\paragraph{Experimental Setup.} To demonstrate the capabilities for super-network generation of BootstrapNAS (and posterior super-network training and searching), we used the already well-optimized models from \cite{huy_phan_2021}, a repository which stores popular models that have been efficiently trained with CIFAR-10 \cite{cifar10}. We selected ResNet-50 \cite{ResnetsHe2016} and MobilenetV2 \cite{MobileNetV2} for BootstrapNAS to generate the corresponding super-networks, train them, and then search for outperforming subnetworks. %In our experiments,
For the search stage, 
BootstrapNAS used NSGA-II with a population size of 50, crossover rate of 0.9, and a mutation rate of 0.02 to search for a Pareto-optimal set.

\paragraph{Results.} BootstrapNAS successfully converted the pre-trained models into super-networks. %We then triggered BootstrapNAS' search capabilities to find Pareto-optimal subnetworks in the model space. 
As Illustrated by Figure \ref{fig:nsga_all}, NSGA-II successfully discovered outstanding subnetworks after 3,000 subnetwork evaluations. These subnetworks outperformed the original pre-trained model in both objectives (MACs and accuracy). 

As Illustrated in Figure \ref{fig:nsga_all}, BootstrapNAS' ResNet-50 super-network contains a myriad of subnetworks that outperform the given pre-trained model. For instance, BootstrapNAS B-RC requires $\sim2.81\times$ fewer MACs than the given pre-trained model while slightly improving the top 1 accuracy (from 93.65\% to 93.70\%). If a small drop in accuracy of $\sim1\%$ is allowed, BootstrapNAS discovers subnetworks, e.g., BootstrapNAS A-RC, that require $\sim3.65\times$ fewer MACs than the original pre-trained model.

In the case of BootstrapNAS' MobilenetV2 super-network, NSGA-II produces a Pareto front with several subnetworks that outperform the given pre-trained model, e.g., BootstrapNAS B-MC, requires $\sim2.39\times$ fewer MACs than the pre-trained model while maintaining its top 1 accuracy ($\sim93.91$\%). If a small drop in accuracy of $\sim1\%$ is allowed, BootstrapNAS discovers subnetworks, e.g. BootstrapNAS A-MC, that require $\sim3.56\times$ fewer MACs than the pre-trained model.

% \todo[inline]{Should we also use Resnet-50 from Imagenet in this paper?}

\section{Conclusion}

%In this paper, we present 
BootstrapNAS %, 
is 
a software framework within NNCF for automatic generation of NAS super-networks.
BootstrapNAS takes as input a pre-trained model, analyzes its architecture, converts it into a super-network, applies %. Once the super-network is available, a 
state-of-the-art techniques 
for training the super-network, and then automatically discovers outperforming subnetworks. 
 
Currently, BootstrapNAS' %can convert arbitrary
focus is on 
convolutional neural networks. In the future, we plan to support other domains, e.g., Natural Language Processing (NLP) models. BootstrapNAS is an open-source project that will be released as part of the Neural Network Compression Framework (NNCF). The example super-networks %generated for our experiments 
presented in this document
are available for reproducibility of the results.

\bibliography{biblio}

\end{document}